\documentclass{article}

\usepackage[final]{nips_2017}

\usepackage[utf8]{inputenc} 
\usepackage[T1]{fontenc}    
\usepackage{hyperref}       
\usepackage{url}            
\usepackage{booktabs}       
\usepackage{amsfonts}       
\usepackage{nicefrac}       
\usepackage{microtype}      
\usepackage{color}
\usepackage{mdwlist}
\usepackage{bm}
\usepackage{amsmath}
\usepackage{graphicx} 
\usepackage{algorithm}
\usepackage{algorithmic}
\usepackage{wrapfig}
\usepackage{url}
\title{Towards Accurate Binary Convolutional Neural Network}

\author{
Xiaofan Lin \ \ \ \ \ \ \ \ \ \  \ \ \ \ \ \ Cong Zhao \ \ \ \ \ \ \ \ \ \  \ \ \ \ \ \ Wei Pan*\\
DJI Innovations Inc, Shenzhen, China \\
\texttt{\{xiaofan.lin, cong.zhao, wei.pan\}@dji.com}
}

\begin{document}

\maketitle
\let\thefootnote\relax\footnotetext{$*$ indicates corresponding author.}

\begin{abstract}
We introduce a novel scheme to train binary convolutional neural networks (CNNs) -- CNNs with weights and activations constrained to \{-1,+1\} at run-time. It has been known that using binary weights and activations drastically reduce memory size and accesses, and can replace arithmetic operations with more efficient bitwise operations, leading to much faster test-time inference and lower power consumption. However, previous works on binarizing CNNs usually result in severe prediction accuracy degradation. In this paper, we address this issue with two major innovations: (1) approximating full-precision weights with the linear combination of multiple binary weight bases; (2) employing multiple binary activations to alleviate information loss. The implementation of the resulting binary CNN, denoted as ABC-Net, is shown to achieve much closer performance to its full-precision counterpart, and even reach the comparable prediction accuracy on ImageNet and forest trail datasets, given adequate binary weight bases and activations.
\end{abstract}

\section{Introduction}
Convolutional neural networks (CNNs) have achieved state-of-the-art results on real-world applications such as image classification \citep{he2016deep} and object detection \citep{ren2015faster}, with the best results obtained with large models and sufficient computation resources. Concurrent to these progresses, the deployment of CNNs on mobile devices for consumer applications is gaining more and more attention, due to the widespread commercial value and the exciting prospect.

On mobile applications, it is typically assumed that training is performed on the server and test or inference is executed on the mobile devices 
\citep{courbariaux2016binarized,esser2016convolutional}.
In the training phase, GPUs enabled substantial breakthroughs because of their greater computational speed. In the test phase, however, GPUs are usually too expensive to deploy. Thus improving the test-time performance and reducing hardware costs are likely to be crucial for further progress, as mobile applications usually require real-time, low power consumption and fully embeddable. As a result, there is much interest in research and development of dedicated hardware for deep neural networks (DNNs). Binary neural networks (BNNs) \citep{courbariaux2016binarized,rastegari2016xnor}, i.e., neural networks with weights and perhaps activations constrained to only two possible values (e.g., -1 or +1), would bring great benefits to specialized DNN hardware for three major reasons: (1) the binary weights/activations reduce memory usage and model size 32 times compared to single-precision version; (2) if weights are binary, then most multiply-accumulate operations can be replaced by simple accumulations, which is beneficial because multipliers are the most space and power-hungry components of the digital implementation of neural networks; (3) furthermore, if both activations and weights are binary, the multiply-accumulations can be replaced by the bitwise operations: xnor and bitcount \cite{courbariaux2016binarized}. This could have a big impact on dedicated deep learning hardware. For instance, a 32-bit floating point multiplier costs about 200 Xilinx FPGA slices \citep{govindu2004analysis},  whereas a 1-bit xnor gate only costs a single slice. Semiconductor manufacturers like IBM \citep{esser2016convolutional} and Intel \citep{venkatesh2016accelerating} have been involved in the research and development of related chips. 

However, binarization usually cause severe prediction accuracy degradation, especially on complex tasks such as classification on ImageNet dataset. To take a closer look, \cite{rastegari2016xnor} shows that binarizing weights causes the accuracy of Resnet-18 drops from 69.3\% to 60.8\% on ImageNet dataset. If further binarize activations, the accuracy drops to 51.2\%. Similar phenomenon can also be found in literatures such as \citep{hubara2016quantized}. Clearly there is a considerable gap between the accuracy of a full-precision model and a binary model.

This paper proposes a novel scheme for binarizing CNNs, which aims to alleviate, or even eliminate the accuracy degradation, while still significantly reducing inference time, resource requirement and power consumption. The paper makes the following major contributions.
\begin{itemize}

\item We approximate full-precision weights with the linear combination of multiple binary weight bases. The weights values of CNNs are constrained to $\{-1,+1\}$, which means convolutions can be implemented by only addition and subtraction (without multiplication), or bitwise operation when activations are binary as well. We demonstrate that 3$\sim$5 binary weight bases are adequate to well approximate the full-precision weights.

\item We introduce multiple binary activations. Previous works have shown that the quantization of activations, especially binarization, is more difficult than that of weights \citep{cai2017deep,courbariaux2016binarized}. By employing five binary activations, we have been able to reduce the Top-1 and Top-5 accuracy degradation caused by binarization to around 5\% on ImageNet compared to the full precision counterpart.
\end{itemize}

It is worth noting that the multiple binary weight bases/activations scheme is preferable to the fixed-point quantization in previous works. In those fixed-point quantized networks one still needs to employ arithmetic operations, such as multiplication and addition, on fixed-point values. Even though faster than floating point, they still require relatively complex logic and can consume a lot of power. Detailed discussions can be found in Section \ref{sect_advanage}.

Ideally, combining more binary weight bases and activations always leads to better accuracy and will eventually get very close to that of full-precision networks. We verify this on ImageNet using Resnet network topology. {\bf This is the first time a binary neural network achieves prediction accuracy comparable to its full-precision counterpart on ImageNet.}

\section{Related work}
\label{gen_inst}
{\bf Quantized Neural Networks:} High precision parameters are not very necessary to reach high performance in deep neural networks. Recent research efforts (e.g., \citep{hubara2016quantized}) have considerably reduced a large amounts of memory requirement and computation complexity by using low bitwidth weights and activations. \citet{zhou2016dorefa} further generalized these schemes and proposed to train CNNs with low bitwidth gradients. By performing the quantization after network training or using the ``straight-through estimator (STE)" \citep{bengio2013estimating}, these works avoided the issues of non-differentiable optimization. While some of these methods have produced good results on datasets such as CIFAR-10 and SVHN, none has produced low precision networks competitive with full-precision models on large-scale classification tasks, such as ImageNet. In fact, \citep{zhou2016dorefa} and \citep{hubara2016quantized} experiment with different combinations of bitwidth for weights and activations, and show that the performance of their highly quantized networks deteriorates rapidly when the weights and activations are quantized to less than 4-bit numbers. \cite{cai2017deep} enhance the performance of a low bitwidth model by addressing the gradient mismatch problem, nevertheless there is still much room for improvement.

{\bf Binarized Neural Networks:} The binary representation for deep models is not a new topic. At the emergence of artificial neural networks, inspired biologically, the unit step function has been used as the activation function \citep{toms1990training}. It is known that binary activation can use spiking response for event-based computation and communication (consuming energy only when necessary) and therefore is energy-efficient \citep{esser2016convolutional}. Recently, \citet{courbariaux2016binarized} introduce Binarized-Neural-Networks (BNNs), neural networks with binary weights and activations at run-time. Different from their work, \citet{rastegari2016xnor} introduce simple, efficient, and accurate approximations to CNNs by binarizing the weights and even the intermediate representations in CNNs. All these works drastically reduce memory consumption, and replace most arithmetic operations with bitwise operations, which potentially lead to a substantial increase in power efficiency.

In all above mentioned works, binarization significantly reduces accuracy. Our experimental results on ImageNet show that we are close to filling the gap between the accuracy of a binary model and its full-precision counterpart. We relied on the idea of finding the best approximation of full-precision convolution using multiple binary operations, and employing multiple binary activations to allow more information passing through.

\section{Binarization methods}
In this section, we detail our binarization method, which is termed ABC-Net (\textbf{A}ccurate-\textbf{B}inary-\textbf{C}onvolutional) for convenience. Bear in mind that during training, the real-valued weights are reserved and updated at every epoch, while in test-time only binary weights are used in convolution.

\subsection{Weight approximation}
\label{subsect_weight_approx}
Consider a $L$-layer CNN architecture. Without loss of generality, we assume the weights of each convolutional layer are tensors of dimension $(w, h, c_{\rm in}, c_{\rm out})$, which represents filter width, filter height, input-channel and output-channel respectively. We propose two variations of binarization method for weights at each layer: 1) approximate weights as a whole and 2) approximate weights channel-wise.

\subsubsection{Approximate weights as a whole}
At each layer, in order to constrain a CNN to have binary weights, we estimate the real-value weight filter ${\bm W} \in \mathbb{R}^{w\times h\times c_{\rm in}\times c_{\rm out}}$ using the linear combination of $M$ binary filters ${\bm B}_1, {\bm B}_2, \cdots, {\bm B}_M \in \{-1,+1\}^{w\times h\times c_{\rm in}\times c_{\rm out}}$ such that ${\bm W} \approx \alpha_1{\bm B}_1 + \alpha_2{\bm B}_2 + \cdots + \alpha_M{\bm B}_M$. To find an optimal estimation, a straightforward way is to solve the following optimization problem:
\begin{equation}
\label{ls_problem}
\begin{split}
\min\limits_{{\bm \alpha}, {\bm B}} &J({\bm \alpha}, {\bm B}) = ||{\bm w} - {\bm B}{\bm \alpha}||^2, \quad \text{s.t.}\ {\bm B}_{ij} \in \{-1, +1\},
\end{split}
\end{equation}
where ${\bm B} = [{\rm vec}({\bm B}_1),{\rm vec}({\bm B}_2),\cdots,{\rm vec}({\bm B}_M)]$, ${\bm w}={\rm vec}({\bm W})$ and ${\bm \alpha} = [\alpha_1, \alpha_2,\cdots,\alpha_M]^{\rm T}$. Here the notation ${\rm vec}(\cdot)$ refers to vectorization.

Although a local minimum solution to \eqref{ls_problem} can be obtained by numerical methods, one could not backpropagate through it to update the real-value weight filter ${\bm W}$. To address this issue, assuming the mean and standard deviation of ${\bm W}$ are ${\rm mean}({\bm W})$ and ${\rm std}({\bm W})$ respectively, we fix ${\bm B}_i$'s as follows:
\begin{equation}
\label{B_Fu_sign}
{\bm B}_i = F_{u_i}({\bm W}):= {\rm sign}({\bm {\bar W}} + u_i{\rm std}({\bm W})), i=1,2,\cdots, M,
\end{equation}
where
$
{\bm {\bar W}} = {\bm W} - {\rm mean}({\bm W}),
$
and $u_i$ is a shift parameter. For example, one can choose $u_i$'s to be
$
\label{ui_fixed}
u_i = - 1 + (i-1)\frac{2}{M-1}, i=1,2,\cdots, M,
$
to shift evenly over the range $\left[- {\rm std}({\bm W}), {\rm std}({\bm W})\right]$, or leave it to be trained by the network. This is based on the observation that the full-precision weights tend to have a symmetric, non-sparse distribution, which is close to Gaussian.  
To gain more intuition and illustrate the approximation effectiveness, an example is visualized in Section \ref{supp:sec:weight} of the supplementary material.

With ${\bm B}_i$'s chosen, \eqref{ls_problem} becomes a linear regression problem
\begin{equation}
\label{ls_problem_B-fixed}
\min\limits_{{\bm \alpha}} J({\bm \alpha}) = ||{\bm w} - {\bm B}{\bm \alpha}||^2,
\end{equation}
in which ${\bm B}_i$'s serve as the bases in the design/dictionary matrix. We can then back-propagate through ${\bm B}_i$'s using the ``straight-through estimator'' (STE) \citep{bengio2013estimating}. Assume $c$ as the cost function, ${\bm A}$ and ${\bm O}$ as the input and output tensor of a convolution respectively, the forward and backward approach of an approximated convolution during training can be computed as follows:
\begin{alignat}{2}
\text{Forward:    } &{\bm B}_{1}, {\bm B}_{2}, \cdots, {\bm B}_{M} = F_{u_1}({\bm W}), F_{u_2}({\bm W}), \cdots, F_{u_M}({\bm W}),\\
& \textrm{Solve \eqref{ls_problem_B-fixed} for ${\bm \alpha}$},\\
    &{\bm O} = \sum \limits_{m=1}^M \alpha_m {\rm Conv}({\bm B}_{m}, {\bm A}) \label{def_binconv}. \\
 \text{Backward:    } &\frac{\partial c}{\partial {\bm W}} = \frac{\partial c}{\partial {\bm O}} \left( \sum_{m=1}^M \alpha_m \frac{\partial {\bm O}}{\partial {\bm B}_{m}} \frac{\partial {\bm B}_{m}}{\partial {\bm W}} \right) 
\stackrel{\text{STE}}{=} \frac{\partial c}{\partial {\bm O}} \left( \sum_{m=1}^M \alpha_m \frac{\partial {\bm O}}{\partial {\bm B}_{m}} \right)
=\sum_{m=1}^M \alpha_m \frac{\partial c}{\partial {\bm B}_{m}}. \label{bp_weights}
\end{alignat}

\begin{figure*}[htp!]
\centering
\includegraphics[scale = 0.45]{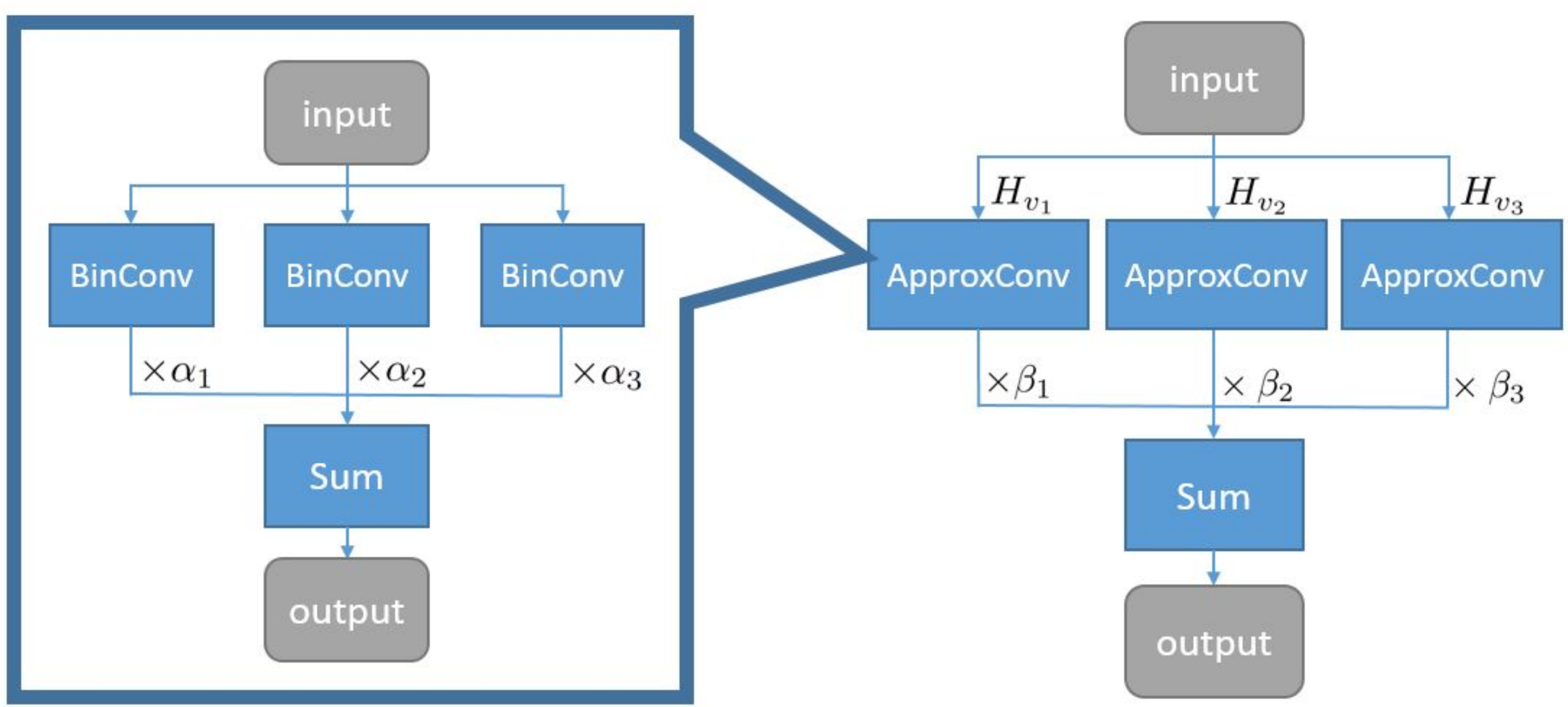}
\caption{An example of the block structure of the convolution in ABC-Net. $M=N=3$. On the left is the structure of the approximated convolution (ApproxConv). ApproxConv is expected to approximate the conventional full-precision convolution with linear combination of binary convolutions (BinConv), i.e., convolution with binary and weights. On the right is the overall block structure of the convolution in ABC-Net. The input is binarized using different functions $H_{v_1}, H_{v_2},H_{v_3}$, passed into the corresponding ApproxConv's and then summed up after multiplying their corresponding $\beta_n$'s. With the input binarized, the BinConv's can be implemented with highly efficient bitwise operations. There are 9 BinConv's in this example and they can work in parallel.}
\label{fig_approx_conv}
\end{figure*}

In test-time, only \eqref{def_binconv} is required. The block structure of this approximated convolution layer is shown on the left side in Figure \ref{fig_approx_conv}. With suitable hardwares and appropriate implementations, the convolution can be efficiently computed. For example, since the weight values are binary, we can implement the convolution with additions and subtractions (thus without multiplications). Furthermore, if the input ${\bm A}$ is binary as well, we can implement the convolution with bitwise operations: xnor and bitcount \citep{rastegari2016xnor}. Note that the convolution with each binary filter can be computed in parallel.
\vspace{-.1in}
\subsubsection{Approximate weights channel-wise}
Alternatively, we can estimate the real-value weight filter ${\bm W}_i \in \mathbb{R}^{w\times h\times c_{\rm in}}$ of each output channel $i \in \{1,2,\cdots, c_{\rm out}\}$ using the linear combination of $M$ binary filters ${\bm B}_{i1}, {\bm B}_{i2}, \cdots, {\bm B}_{iM} \in \{-1,+1\}^{w\times h\times c_{\rm in}}$ such that ${\bm W}_i \approx \alpha_{i1}{\bm B}_{i1} + \alpha_{i2}{\bm B}_{i2} + \cdots + \alpha_{iM}{\bm B}_{iM}$. Again, to find an optimal estimation, we solve a linear regression problem analogy to \eqref{ls_problem_B-fixed} for each output channel. After convolution, the results are concatenated together along the output-channel dimension. If $M=1$, this approach reduces to the Binary-Weights-Networks (BWN) proposed in \citep{rastegari2016xnor}.

Compared to weights approximation as a whole, the channel-wise approach approximates weights more elaborately, however no extra cost is needed during inference. Since this approach requires more computational resources during training, we leave it as a future work and focus on the former approximation approach in this paper.

\subsection{Multiple binary activations and bitwise convolution}
As mentioned above, a convolution can be implemented without multiplications when weights are binarized. However, to utilize the bitwise operation, the activations must be binarized as well, as they are the inputs of convolutions.

Similar to the activation binarization procedure in \citep{zhou2016dorefa}, we binarize activations after passing it through a bounded activation function $h$, which ensures $h(x)\in [0,1]$. We choose the bounded rectifier as $h$. Formally, it can be defined as:
\vspace{-0.1 cm}
\begin{equation}
\label{hv_clip}
h_v(x) = {\rm clip}(x+v, 0, 1),
\end{equation}
where $v$ is a shift parameter. If $v=0$, then $h_v$ is the clip activation function in \citep{zhou2016dorefa}.

We constrain the binary activations to either 1 or -1. In order to transform the real-valued activation ${\bm R}$ into binary activation, we use the following binarization function:
\begin{equation}
H_v({\bm R}):= 2\mathbb{I}_{h_v({\bm R}) \ge 0.5}-1,
\end{equation}
where $\mathbb{I}$ is the indicator function. The conventional forward and backward approach of the  activation can be given as follows: 
\begin{equation}
\begin{aligned}
\text{Forward:    } & {\bm A} = H_v({\bm R}). \\
\text{Backward:   } &\frac{\partial c}{\partial {\bm R}} = \frac{\partial c}{\partial {\bm A}} \circ \mathbb{I}_{0\le {\bm R}-v\le 1}.\quad {\rm (using\ STE)}
\end{aligned}
\end{equation}

Here $\circ$ denotes the Hadamard product. As can be expected, binaizing activations as above is kind of crude and leads to non-trivial losses in accuracy, as shown in \cite{rastegari2016xnor,hubara2016quantized}. While it is also possible to approximate activations with linear regression, as that of weights, another critical challenge arises -- unlike weights, the activations always vary in test-time inference. Luckily, this difficulty can be avoided by exploiting the statistical structure of the activations of deep networks. 

Our scheme can be described as follows. First of all, to keep the distribution of activations relatively stable, we resort to batch normalization \citep{ioffe2015batch}. This is a widely used normalization technique, which forces the responses of each network layer to have zero mean and unit variance. We apply this normalization before activation. Secondly, we estimate the real-value activation ${\bm R}$ using the linear combination of $N$ binary activations ${\bm A}_1, {\bm A}_2, \cdots, {\bm A}_N$ such that ${\bm R} \approx \beta_1 {\bm A}_1 + \beta_2 {\bm A}_2 + \cdots + \beta_N {\bm A}_N$, where 
\begin{equation}
\label{def_multi_act}
{\bm A}_1, {\bm A}_2, \cdots, {\bm A}_N = H_{v_1}({\bm R}),H_{v_2}({\bm R}),\cdots,H_{v_N}({\bm R}).
\end{equation}
Different from that of weights, the parameters $\beta_n$'s and $v_n$'s ($n=1,\cdots,N$) here are both trainable, just like the scale and shift parameters in batch normalization. Without the explicit linear regression approach, $\beta_n$'s and $v_n$'s are tuned by the network itself during training and fixed in test-time. They are expected to learn and utilize the statistical features of full-precision activations.

The resulting network architecture outputs multiple binary activations ${\bm A}_1, {\bm A}_2, \cdots, {\bm A}_N$ and their corresponding coefficients $\beta_1, \beta_2, \cdots, \beta_N$, which allows more information passing through compared to the former one. Combining with the weight approximation, the whole convolution scheme is given by:
\begin{equation}
\label{def_bitconv}
{\rm Conv}({\bm W}, {\bm R}) \approx {\rm Conv}\left(\sum_{m=1}^M \alpha_m{\bm B}_m, \sum_{n=1}^N \beta_n {\bm A}_n \right) \\
= \sum_{m=1}^M \sum_{n=1}^N \alpha_m \beta_n {\rm Conv}\left({\bm B}_m, {\bm A}_n \right),
\end{equation}
which suggests that it can be implemented by computing $M\times N$ bitwise convolutions in parallel. An example of the whole convolution scheme is shown in Figure \ref{fig_approx_conv}.

\subsection{Training algorithm}
\label{sec:algorithm}
A typical block in CNN contains several different layers, which are usually in the following order: (1) Convolution, (2) Batch Normalization, (3) Activation and (4) Pooling. The batch normalization layer \citep{ioffe2015batch} normalizes the input batch by its mean and variance. The activation is an element-wise non-linear function (e.g., Sigmoid, ReLU). The pooling layer applies any type of pooling (e.g., max,min or average) on the input batch. In our experiment, we observe that applying max-pooling on binary input returns a tensor that most of its elements are equal to +1, resulting in a noticeable drop in accuracy. Similar phenomenon has been reported in \cite{rastegari2016xnor} as well. Therefore, we put the max-pooling layer before the batch normalization and activation.

Since our binarization scheme approximates full-precision weights, using the full-precision pre-train model serves as a perfect initialization. However, fine-tuning is always required for the weights to adapt to the new network structure. The training procedure, i.e., ABC-Net, is summarized in Section~\ref{sec:supp:algorithm} of the supplementary material.

It is worth noting that as $M$ increases, the shift parameters get closer and the bases of the linear combination are more correlated, which sometimes lead to rank deficiency when solving \eqref{ls_problem_B-fixed}. This can be tackled with the $\ell_2$ regularization. 

\vspace*{-0.2 cm}
\section{Experiment results}
In this section, the proposed ABC-Net was evaluated on the ILSVRC12 ImageNet classification dataset \citep{deng2009imagenet}, and  visual perception of forest trails datasets for mobile robots \citep{giusti2016machine} in Section \ref{sec:supp:forest} of supplementary material.

\subsection{Experiment results on ImageNet dataset}
The ImageNet dataset contains about 1.2 million high-resolution natural images for training that spans 1000 categories of objects. The validation set contains 50k images. We use Resnet (\citep{he2016deep}) as network topology. The images are resized to 224x224 before fed into the network. We report our classification performance using Top-1 and Top-5 accuracies. 

\subsubsection{Effect of weight approximation}
We first evaluate the weight approximation technique by examining the accuracy improvement for a binary model. To eliminate variables, we leave the activations being full-precision in this experiment. Table \ref{table-accuracy-fpact} shows the prediction accuracy of ABC-Net on ImageNet with different choices of $M$. For comparison, we add the results of Binary-Weights-Network (denoted `BWN') reported in \cite{rastegari2016xnor} and the full-precision network (denoted `FP'). The BWN binarizes weights and leaves the activations being full-precision as we do. All results in this experiment use Resnet-18 as network topology. It can be observed that as $M$ increases, the accuracy of ABC-Net converges to its full-precision counterpart. The Top-1 gap between them reduces to only 0.9 percentage point when $M=5$, which suggests that this approach nearly eliminates the accuracy degradation caused by binarizing weights.
{\small
\begin{table*}[!ht]
\caption{Top-1 (left) and Top-5 (right) accuracy of ABC-Net on ImageNet, using full-precision activation and different choices of the number of binary weight bases $M$.}
\label{table-accuracy-fpact}
\begin{center}
\begin{tabular}{c|c|c|c|c|c|c}
\hline 
\multicolumn{1}{c|}{} &\multicolumn{1}{c|}{${\rm BWN}$} &\multicolumn{1}{c|}{$M=1$} &\multicolumn{1}{c|}{$M=2$} &\multicolumn{1}{c|}{$M=3$} &\multicolumn{1}{c|}{$M=5$} &\multicolumn{1}{c}{${\rm FP}$} 
\\ \hline \hline 
Top-1 & 60.8\% & 62.8\% & 63.7\% & 66.2\%  &  68.3\% & 69.3\% \\
\hline
Top-5 & 83.0\% & 84.4\% & 85.2\% & 86.7\%  &  87.9\% & 89.2\% \\ 
\hline
\end{tabular}
\end{center}
\end{table*}
}

\vspace*{-0.3 cm}
For interested readers, Figure \ref{fig_bwn_acc} in section~\ref{supp:sec:bwn_acc} of the supplementary material shows that the relationship between accuracy and $M$ appears to be linear. Also, in Section~\ref{supp:sec:weight} of the supplementary material, a visualization of the approximated weights is provided.

\subsubsection{Configuration space exploration}
\label{subsect_space_exp}
We explore the configuration space of combinations of number of weight bases and activations. Table \ref{lossless-BNN-accuracy-table} presents the results of ABC-Net with different configurations. The parameter settings for these experiments are provided in Section \ref{supp:sec:param} of the supplementary material.
\vspace{-0.1 cm}
{\small
\begin{table*}[!ht]
\caption{
Prediction accuracy (Top-1/Top-5) for ImageNet with different choices of $M$ and $N$ in a ABC-Net (approximate weights as a whole). ``res18'', ``res34'' and ``res50'' are short for Resnet-18, Resnet-34 and Resnet-50 network topology respectively. $M$ and $N$ refer to the number of weight bases and activations respectively. 
}
\label{lossless-BNN-accuracy-table}
\begin{center}
\begin{tabular}{c|c|c|c|c|c|c}
\hline 
\multicolumn{1}{c|}{Network} &\multicolumn{1}{c|}{$M$-weight base} &\multicolumn{1}{c|}{$N$-activation base} &\multicolumn{1}{c|}{Top-1} &\multicolumn{1}{c|}{Top-5} &\multicolumn{1}{c|}{Top-1 gap} &\multicolumn{1}{c}{Top-5 gap}
\\ \hline \hline 
res18 & {1} &  1   & 42.7\% & 67.6\% & 26.6\% & 21.6\%\\ 
\hline \hline 
res18 & {3} &  1   & 49.1\% & 73.8\% & 20.2\% & 15.4\%\\
res18 & {3} &  3   & 61.0\% & 83.2\% & 8.3\%  & 6.0\%\\
res18 & {3} &  5   & 63.1\% & 84.8\% & 6.2\%  & 4.4\%\\ 
\hline \hline 
res18 & {5} &  1   & 54.1\% & 78.1\% & 15.2\% & 11.1\%\\
res18 & {5} &  3   & 62.5\% & 84.2\% & 6.8\%  & 5.0\% \\
res18 & {5} &  5   & 65.0\% & 85.9\% & {\bf 4.3\%} & {\bf 3.3}\%\\
\hline \hline 
res18 & \multicolumn{2}{c|}{Full Precision} & 69.3\% & 89.2\% & - & - \\
\hline \hline
res34 & {1} &  1   & 52.4\% & 76.5\% & 20.9\% & 14.8\% \\
res34 & {3} &  3   & 66.7\% & 87.4\% & 6.6\% & 3.9\% \\
res34 & {5} &  5   & 68.4\% & 88.2\% & {\bf 4.9\%} & {\bf  3.1}\%\\
\hline\hline
res34 & \multicolumn{2}{c|}{Full Precision} & 73.3\% & 91.3\% & - & -\\
\hline\hline
res50 & {5} &  5   & 70.1\% & 89.7\% & {\bf 6.0\%} & {\bf  3.1}\%\\
\hline
res50 & \multicolumn{2}{c|}{Full Precision} & 76.1\% & 92.8\% & - & -\\
\hline
\end{tabular}
\end{center}
\end{table*}
}
\vspace*{-0.06 cm}

As balancing between multiple factors like training time and inference time, model size and accuracy is more a problem of practical trade-off, there will be no definite conclusion as which combination of ($M, N$) one should choose. In general, Table \ref{lossless-BNN-accuracy-table} shows that (1) the prediction accuracy of ABC-Net improves greatly as the number of binary activations increases, which is analogous to the weight approximation approach; (2) larger $M$ or $N$ gives better accuracy; (3) when $M=N=5$, the Top-1 gap between the accuracy of a full-precision model and a binary one reduces to around 5\%. To gain a visual understanding and show the possibility of extensions to other tasks such object detection, we print the a sample of feature maps in Section \ref{sec:supp:featuremap} of supplementary material.

\subsubsection{Comparison with the state-of-the-art}
{\small
\begin{table*}[!htp]
\caption{Classification test accuracy of CNNs trained on ImageNet with Resnet-18 network topology. `W' and `A' refer to the weight and activation bitwidth respectively.}
\label{imagenet-table}
\begin{center}
\begin{tabular}{c|c|c|c|c}
\hline 
\multicolumn{1}{c|}{Model}   &\multicolumn{1}{c|}{W} &\multicolumn{1}{c|}{A} &\multicolumn{1}{c|}{Top-1} &\multicolumn{1}{c}{Top-5}
\\ \hline \hline 
{Full-Precision Resnet-18 [full-precision weights and activation]}      &        32     &       32   & 69.3\%    &   89.2\%   \\
{BWN [full-precision activation] \cite{rastegari2016xnor} }& 1&32   & 60.8\%    &   83.0\%   \\
{DoReFa-Net [1-bit weight and 4-bit activation] \cite{zhou2016dorefa} }&   1    & 4    & 59.2\%    &   81.5\%   \\ 
{XNOR-Net [binary weight and activation] \cite{rastegari2016xnor}}    &        1      &1    & 51.2\%    &   73.2\%   \\
{BNN [binary weight and activation] \cite{courbariaux2016binarized}}       &      1           &  1    & 42.2\%    & 67.1\%     \\
\hline \hline 
{ABC-Net [5 binary weight bases, 5 binary activations]}       &      1           &  1   & 65.0\% & 85.9\% \\
{ABC-Net [5 binary weight bases, full-precision activations]}       &      1         & 32    & 68.3\%  & 87.9\% \\
\hline
\end{tabular}
\end{center}
\end{table*}
}

Table \ref{imagenet-table} presents a comparison between ABC-Net and several other state-of-the-art models, i.e., full-precision Resnet-18, BWN and XNOR-Net in \citep{rastegari2016xnor}, DoReFa-Net in \citep{zhou2016dorefa} and BNN in \citep{courbariaux2016binarized}\footnote{\citet{courbariaux2016binarized} did not report their result on ImageNet. We implemented and presented the result.} respectively. All comparative models use Resnet-18 as network topology. The full-precision Resnet-18 achieves 69.3\% Top-1 accuracy. Although \citet{rastegari2016xnor}'s BWN model and DeReFa-Net perform well, it should be noted that they use full-precision and 4-bit activation respectively. Models (XNOR-Net and BNN) that used both binary weights and activations achieve much less satisfactory accuracy, and is significantly outperformed by ABC-Net with multiple binary weight bases and activations. It can be seen that ABC-Net has achieved state-of-the-art performance as a binary model.

One might argue that 5-bit width quantization scheme could reach similar accuracy as that of ABC-Net with 5 weight bases and 5 binary activations. However, the former one is less efficient and requires distinctly more hardware resource. More detailed discussions can be found in Section \ref{sect_advanage}.
\vspace*{-0.3 cm}

\section{Discussion}
\subsection{Why adding a shift parameter works?}
Intuitively, the multiple binarized weight bases/activations scheme works because it allows more information passing through. Consider the case that a real value, say 1.5, is passed to a binarized function $f(x)={\rm sign}(x)$, where ${\rm sign}$ maps a positive $x$ to 1 and otherwise -1. In that case, the outputs of $f(1.5)$ is 1, which suggests that the input value is positive. Now imagine that we have two binarization function $f_1(x)={\rm sign}(x)$ and $f_2(x)={\rm sign}(x-2)$. In that case $f_1$ outputs 1 and $f_2$ outputs -1, which suggests that the input value is not only positive, but also must be smaller than 2. Clearly we see that each function contributes differently to represent the input and more information is gained from $f_2$ compared to the former case.

From another point of view, both coefficients ($\beta$'s) and shift parameters are expected to learn and utilize the statistical features of full-precision tensors, just like the scale and shift parameters in batch normalization. If we have more binarized weight bases/activations, the network has the capacity to approximate the full-precision one more precisely. Therefore, it can be deduced that when $M$ or $N$ is large enough, the network learns to tune itself so that the combination of $M$ weight bases or $N$ binarized activations can act like the full-precision one.

\subsection{Advantage over the fixed-point quantization scheme}
\label{sect_advanage}
It should be noted that there are key differences between the multiple binarization scheme ($M$ binarized weight bases or $N$ binarized activations) proposed in this paper and the fixed-point quantization scheme in the previous works such as \citep{zhou2016dorefa,hubara2016quantized}, though at first thought $K$-bit width quantization seems to share the same memory requirement with $K$ binarizations. Specifically, our $K$ binarized weight bases/activations is preferable to the fixed K-bit width scheme for the following reasons: 

(1) The $K$ binarization scheme preserves binarization for bitwise operations. One or several bitwise operations is known to be more efficient than a fixed-point multiplication, which is a major reason that BNN/XNOR-Net was proposed. 

(2) A $K$-bit width multiplier consumes more resources than $K$ 1-bit multipliers in a digital chip: it requires more than $K$ bits to store and compute, otherwise it could easily overflow/underflow. For example, if a real number is quantized to a 2-bit number, a possible choice is in range \{0,1,2,4\}. In this 2-bit multiplication, when both numbers are 4, it outputs $4\times 4=16$, which is not within the range. In \citep{zhou2016dorefa}, the range of activations is constrained within [0,1], which seems to avoid this situation. However, fractional numbers do not solve this problem, severe precision deterioration will appear during the multiplication if there are no extra resources. The fact that the complexity of a multiplier is proportional to THE SQUARE of bit-widths can be found in literatures (e.g., sec 3.1.1. in \citep{grabbe2003fpga}). In contrast, our $K$ binarization scheme does not have this issue -- it always outputs within the range \{-1,1\}. The saved hardware resources can be further used for parallel computing. 

(3) A binary activation can use spiking response for event-based computation and communication (consuming energy only when necessary) and therefore is energy-efficient \citep{esser2016convolutional}. This can be employed in our scheme, but not in the fixed $K$-bit width scheme. Also, we have mentioned the fact that $K$-bit width multiplier consumes more resources than $K$ 1-bit multipliers. It is noteworthy that these resources include power. 

To sum up, $K$-bit multipliers are the most space and power-hungry components of the digital implementation of DNNs. Our scheme could bring great benefits to specialized DNN hardware.

\subsection{Further computation reduction in run-time}
On specialized hardware, the following operations in our scheme can be integrated with other operations in run-time and further reduce the computation requirement.

(1) Shift operations. The existence of shift parameters seem to require extra additions/subtractions (see \eqref{B_Fu_sign} and \eqref{hv_clip}). However, the binarization operation with a shift parameter can be implemented as a comparator where the shift parameter is the number for comparison, e.g., $H_v({\bm R}) = \left\{\begin{matrix}
{\bm 1}, & {\bm R} \ge 0.5-v; \\ 
\bm{-1}, & {\bm R} < 0.5-v.
\end{matrix}\right.$ ($0.5-v$ is a constant), so no extra additions/subtractions are involved.

(2) Batch normalization. In run-time, a batch normalization is simply an affine function, say, ${\rm BN}({\bm R})=a{\bm R}+b$, whose scale and shift parameters $a,b$ are fixed and can be integrated with $v_n$'s. More specifically, a batch normalization can be integrated into a binarization operation as follow: $H_v({\rm BN}({\bm R}))=\left\{\begin{matrix}
{\bm 1}, & a{\bm R}+b \ge 0.5-v; \\ 
\bm{-1}, & a{\bm R}+b < 0.5-v.
\end{matrix}\right. = \left\{\begin{matrix}
{\bm 1}, & {\bm R} \ge (0.5-v-b)/a; \\ 
\bm{-1}, & {\bm R} < (0.5-v-b)/a.
\end{matrix}\right.$ Therefore, there will be no extra cost for the batch normalization.

\section{Conclusion and future work}
We have introduced a novel binarization scheme for weights and activations during forward and backward propagations called ABC-Net. We have shown that it is possible to train a binary CNN with ABC-Net on ImageNet and achieve accuracy close to its full-precision counterpart. The binarization scheme proposed in this work is parallelizable and hardware friendly, and the impact of such a method on specialized hardware implementations of CNNs could be major, by replacing most multiplications in convolution with bitwise operations. The potential to speed-up the test-time inference might be very useful for real-time embedding systems. Future work includes the extension of those results to other tasks such as object detection and other models such as RNN. Also, it would be interesting to investigate using FPGA/ASIC or other customized deep learning processor \citep{liu2016cambricon} to implement ABC-Net at run-time.

\section{Acknowledgement}
We acknowledge Mr Jingyang Xu for helpful discussions.

{\small 
\bibliography{ref}

\begin{thebibliography}{19}
\providecommand{\natexlab}[1]{#1}
\providecommand{\url}[1]{\texttt{#1}}
\expandafter\ifx\csname urlstyle\endcsname\relax
  \providecommand{\doi}[1]{doi: #1}\else
  \providecommand{\doi}{doi: \begingroup \urlstyle{rm}\Url}\fi

\bibitem[Bengio et~al.(2013)Bengio, L{\'e}onard, and
  Courville]{bengio2013estimating}
Y.~Bengio, N.~L{\'e}onard, and A.~Courville.
\newblock Estimating or propagating gradients through stochastic neurons for
  conditional computation.
\newblock \emph{arXiv preprint arXiv:1308.3432}, 2013.

\bibitem[Cai et~al.(2017)Cai, He, Sun, and Vasconcelos]{cai2017deep}
Z.~Cai, X.~He, J.~Sun, and N.~Vasconcelos.
\newblock Deep learning with low precision by half-wave gaussian quantization.
\newblock \emph{arXiv preprint arXiv:1702.00953}, 2017.

\bibitem[Courbariaux et~al.(2016)Courbariaux, Hubara, Soudry, El-Yaniv, and
  Bengio]{courbariaux2016binarized}
M.~Courbariaux, I.~Hubara, D.~Soudry, R.~El-Yaniv, and Y.~Bengio.
\newblock Binarized neural networks: Training deep neural networks with weights
  and activations constrained to+ 1 or-1.
\newblock \emph{arXiv preprint arXiv:1602.02830}, 2016.

\bibitem[Deng et~al.(2009)Deng, Dong, Socher, Li, Li, and
  Fei-Fei]{deng2009imagenet}
J.~Deng, W.~Dong, R.~Socher, L.-J. Li, K.~Li, and L.~Fei-Fei.
\newblock Imagenet: A large-scale hierarchical image database.
\newblock In \emph{Computer Vision and Pattern Recognition, 2009. CVPR 2009.
  IEEE Conference on}, pages 248--255. IEEE, 2009.

\bibitem[Esser et~al.(2016)Esser, Merolla, Arthur, Cassidy, Appuswamy,
  Andreopoulos, Berg, McKinstry, Melano, Barch, et~al.]{esser2016convolutional}
S.~K. Esser, P.~A. Merolla, J.~V. Arthur, A.~S. Cassidy, R.~Appuswamy,
  A.~Andreopoulos, D.~J. Berg, J.~L. McKinstry, T.~Melano, D.~R. Barch, et~al.
\newblock Convolutional networks for fast, energy-efficient neuromorphic
  computing.
\newblock \emph{Proceedings of the National Academy of Sciences}, page
  201604850, 2016.

\bibitem[Giusti et~al.(2016)Giusti, Guzzi, Ciresan, He, Rodriguez, Fontana,
  Faessler, Forster, Schmidhuber, Di~Caro, Scaramuzza, and
  Gambardella]{giusti2016machine}
A.~Giusti, J.~Guzzi, D.~Ciresan, F.-L. He, J.~P. Rodriguez, F.~Fontana,
  M.~Faessler, C.~Forster, J.~Schmidhuber, G.~Di~Caro, D.~Scaramuzza, and
  L.~Gambardella.
\newblock A machine learning approach to visual perception of forest trails for
  mobile robots.
\newblock \emph{IEEE Robotics and Automation Letters}, 2016.

\bibitem[Govindu et~al.(2004)Govindu, Zhuo, Choi, and
  Prasanna]{govindu2004analysis}
G.~Govindu, L.~Zhuo, S.~Choi, and V.~Prasanna.
\newblock Analysis of high-performance floating-point arithmetic on fpgas.
\newblock In \emph{Parallel and Distributed Processing Symposium, 2004.
  Proceedings. 18th International}, page 149. IEEE, 2004.

\bibitem[Grabbe et~al.(2003)Grabbe, Bednara, Teich, von~zur Gathen, and
  Shokrollahi]{grabbe2003fpga}
C.~Grabbe, M.~Bednara, J.~Teich, J.~von~zur Gathen, and J.~Shokrollahi.
\newblock Fpga designs of parallel high performance gf (2233) multipliers.
\newblock In \emph{ISCAS (2)}, pages 268--271. Citeseer, 2003.

\bibitem[He et~al.(2016)He, Zhang, Ren, and Sun]{he2016deep}
K.~He, X.~Zhang, S.~Ren, and J.~Sun.
\newblock Deep residual learning for image recognition.
\newblock In \emph{Proceedings of the IEEE Conference on Computer Vision and
  Pattern Recognition}, pages 770--778, 2016.

\bibitem[Hubara et~al.(2016)Hubara, Courbariaux, Soudry, El-Yaniv, and
  Bengio]{hubara2016quantized}
I.~Hubara, M.~Courbariaux, D.~Soudry, R.~El-Yaniv, and Y.~Bengio.
\newblock Quantized neural networks: Training neural networks with low
  precision weights and activations.
\newblock \emph{arXiv preprint arXiv:1609.07061}, 2016.

\bibitem[Ioffe and Szegedy(2015)]{ioffe2015batch}
S.~Ioffe and C.~Szegedy.
\newblock Batch normalization: Accelerating deep network training by reducing
  internal covariate shift.
\newblock \emph{arXiv preprint arXiv:1502.03167}, 2015.

\bibitem[Kingma and Ba(2014)]{kingma2014adam}
D.~Kingma and J.~Ba.
\newblock Adam: A method for stochastic optimization.
\newblock \emph{arXiv preprint arXiv:1412.6980}, 2014.

\bibitem[Liu et~al.(2016)Liu, Du, Tao, Han, Luo, Xie, Chen, and
  Chen]{liu2016cambricon}
S.~Liu, Z.~Du, J.~Tao, D.~Han, T.~Luo, Y.~Xie, Y.~Chen, and T.~Chen.
\newblock Cambricon: An instruction set architecture for neural networks.
\newblock In \emph{Proceedings of the 43rd International Symposium on Computer
  Architecture}, pages 393--405. IEEE Press, 2016.

\bibitem[Qian(1999)]{qian1999momentum}
N.~Qian.
\newblock On the momentum term in gradient descent learning algorithms.
\newblock \emph{Neural networks}, 12\penalty0 (1):\penalty0 145--151, 1999.

\bibitem[Rastegari et~al.(2016)Rastegari, Ordonez, Redmon, and
  Farhadi]{rastegari2016xnor}
M.~Rastegari, V.~Ordonez, J.~Redmon, and A.~Farhadi.
\newblock Xnor-net: Imagenet classification using binary convolutional neural
  networks.
\newblock In \emph{European Conference on Computer Vision}, pages 525--542.
  Springer, 2016.

\bibitem[Ren et~al.(2015)Ren, He, Girshick, and Sun]{ren2015faster}
S.~Ren, K.~He, R.~Girshick, and J.~Sun.
\newblock Faster r-cnn: Towards real-time object detection with region proposal
  networks.
\newblock In \emph{Advances in neural information processing systems}, pages
  91--99, 2015.

\bibitem[Toms(1990)]{toms1990training}
D.~Toms.
\newblock Training binary node feedforward neural networks by back propagation
  of error.
\newblock \emph{Electronics letters}, 26\penalty0 (21):\penalty0 1745--1746,
  1990.

\bibitem[Venkatesh et~al.(2016)Venkatesh, Nurvitadhi, and
  Marr]{venkatesh2016accelerating}
G.~Venkatesh, E.~Nurvitadhi, and D.~Marr.
\newblock Accelerating deep convolutional networks using low-precision and
  sparsity.
\newblock \emph{arXiv preprint arXiv:1610.00324}, 2016.

\bibitem[Zhou et~al.(2016)Zhou, Wu, Ni, Zhou, Wen, and Zou]{zhou2016dorefa}
S.~Zhou, Y.~Wu, Z.~Ni, X.~Zhou, H.~Wen, and Y.~Zou.
\newblock Dorefa-net: Training low bitwidth convolutional neural networks with
  low bitwidth gradients.
\newblock \emph{arXiv preprint arXiv:1606.06160}, 2016.

\end{thebibliography}
\bibliographystyle{abbrvnat}
}

\newpage

{\large
\begin{center}
{\centering \textbf{Supplementary Material}}
\end{center}
}
\numberwithin{equation}{section}

\makeatletter
\renewcommand{\thesection}{S\arabic{section}}   
\renewcommand{\thesubsection}{S\arabic{subsection}}   
\renewcommand{\thetable}{S\arabic{table}}   
\renewcommand{\thefigure}{S\arabic{figure}}

\numberwithin{equation}{subsection}
\subsection{Summary of training algorithm in Section~\ref{sec:algorithm}}
\label{sec:supp:algorithm}
\begin{algorithm}[!ht] 
\label{alg:train}
\caption{
Training a $L$-layer ABC-Net. $c$ is the cost function for minibatch, and $\lambda$ is the learning rate decay factor. $\circ$ indicates element-wise multiplication. {\rm BatchNorm}() specifies how to batch-normalize the output of convolution and {\rm BackBatchNorm} specifies how to backpropagate through the normalization \citep{ioffe2015batch}. Analogously, {\rm Conv}() and {\rm BackConv}() respectively specify how to do convolution and how to backpropagate through the convolution. {\rm Update}() specifies how to update the parameters when their gradients are known, such as ADAM \citep{kingma2014adam} and Momentum \citep{qian1999momentum})}
\begin{algorithmic}
    \REQUIRE a minibatch of inputs and targets, 
    number of binary weight bases $M$,
    number of binary activations $N$,
    previous weights ${\bm W}$, 
    and learning rate $\eta$.
    \ENSURE updated weights ${\bm W}$ and updated learning rate $\eta$.
    
    \STATE \COMMENT{1. Computing the parameters gradients:}
    
    \STATE \COMMENT{1.1. Forward propagation:}
    \FOR{$l=1$ to $L$}  
        \STATE Compute $\alpha_m^l, {\bm B}_m^l, m=1,\cdots,M$ with \eqref{B_Fu_sign} and \eqref{ls_problem_B-fixed})
        \STATE $s^l \leftarrow {\rm Conv}({\bm W}^l, {\bm R}^{l-1})$ using \eqref{def_bitconv}
        \STATE {\rm Optionally apply max-pooling}
        \STATE $a^l \leftarrow {\rm BatchNorm}(s^l)$
        \IF{$l < L$}
            \STATE ${\bm A}_n^l \leftarrow H_{v_n}(a^l), n=1,2,\cdots,N$  (using \eqref{def_multi_act})
        \ENDIF
    \ENDFOR
    
    \STATE \COMMENT{1.2. Backward propagation:}
    \STATE \COMMENT{Note that the gradients are full-precision.}
    \STATE Compute $g_{a_L}=\frac{\partial c}{\partial a_L}$
    \FOR{$l=L$ to $1$} 
        \IF{$l < L$}
            \STATE $g_{a^l} \leftarrow \sum_{n=1}^N \beta_n g_{{\bm A}_n^l} \circ \mathbb{I}_{0\le a^l-v_n\le 1}$
        \ENDIF
        \STATE $g_{s_l} \leftarrow {\rm BackBatchNorm}(g_{a^l}, s_l)$
        \STATE $g_{\beta_m}, g_{{\bm B}_m^l} \leftarrow {\rm BackConv}(g_{s_l}, {\bm B}_m^{l-1}, {\bm A}_m^{l-1})$
    \ENDFOR
    
    \STATE \COMMENT{2. Accumulating the parameters gradients:}
    \FOR{$l=1$ to $L$}
        \STATE With $g_{{\bm B}_m^l}$ known, compute $g_{{\bm W}^l}$ using \eqref{bp_weights}
        \STATE ${\bm W}^l \leftarrow {\rm Update}({\bm W}^l, \eta, g_{{\bm W}^l})$
        \STATE $\beta_m \leftarrow {\rm Update}(\beta_m, \eta, g_{\beta_m}), m=1,2,\cdots,M$
        \STATE $v_m \leftarrow {\rm Update}(v_m, \eta, g_{v_m}), m=1,2,\cdots,M$
        \STATE $\eta \leftarrow \lambda \eta$
    \ENDFOR
\end{algorithmic}
\end{algorithm}

\subsection{Weight approximation}
\label{supp:sec:weight}
\begin{figure*}[htb]
\centering
\includegraphics[width=1.0\textwidth]{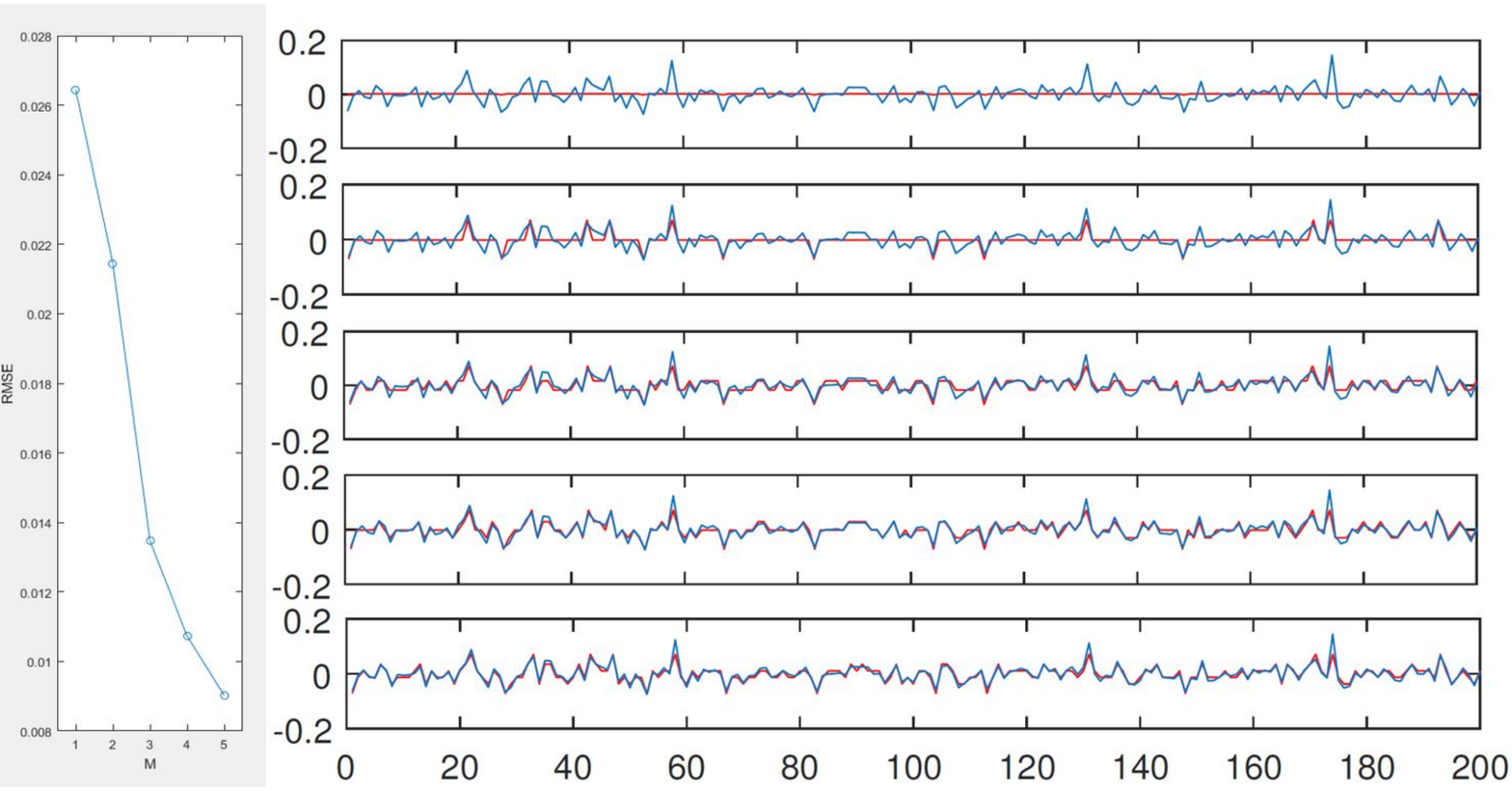}
\caption{Fitting a section of weights of full-precision Resnet-18 trained on ImageNet. On the left side the RMSE for the estimated weights with increasing number of bases is shown, and on the right side 5 fitting results are shown, whose choice of $M$ are respectively 1 to 5 from top to bottom (Blue line: full-precision weights; Red line: estimated weights).}
\label{fig_approx_M}
\end{figure*}

In this section we explore how well the weight approximation can achieved given adequate binary weight bases (Section \ref{subsect_weight_approx}). To gain a visual intuition, we randomly sample a slice of weight tensor from a full-precision Resnet-18 model pretrained on ImageNet. The sliced tensor is then vectorized, and we approximate it with $M$ bases using linear regression (see \eqref{ls_problem_B-fixed}). The results are presented in Figure \ref{fig_approx_M}, where the left subfigure shows the root mean square (RMSE) for the estimated weights with increasing number of bases, and the right one shows 5 fitting results, whose choice of $M$ are respectively 1 to 5 from top to bottom. The blue line in the right subfigure draws the groundtruth weights from the full-precision pretrained model, and the red line is the estimated one. It can be observed that $M=3$ is adequate to have a rough fitting, and it gets almost perfect when $M=5$.

\subsection{Feature map}
\label{sec:supp:featuremap}
\begin{figure*}[htb]
\centering
\includegraphics[width=1.0\textwidth]{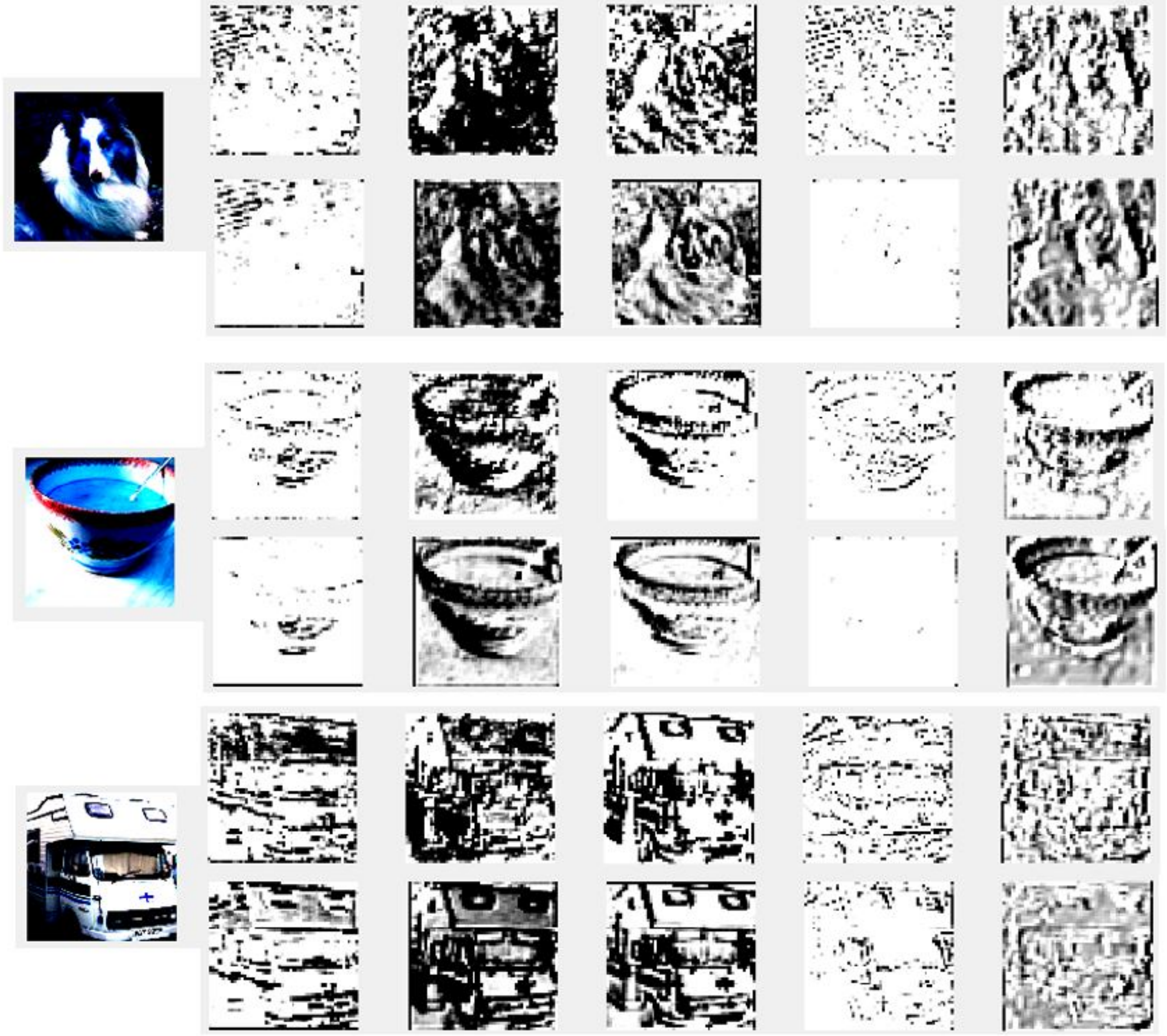}
\caption{Examples of feature maps. The feature maps from the first convolution layer of ABC-Net (above) looks similar to that of its full-precision counterpart (below). Settings for the ABC-Net: $M=5,N=3$, using Resnet-18.}
\label{fig_featmap}
\end{figure*}

It is also possible to perform more complex tasks beyond classification using ABC-Net, as long as the model is built upon a CNN, such as faster RCNN for object detection, in which the classification model serves as a pre-train model. Thus, one might be interested in whether ABC-Net learns similar feature maps as its full-precision counterpart. Figure \ref{fig_featmap} shows several example image and the corresponding feature maps of these two models, from which we see that they are indeed similar. This shows the potential for ABC-Net to further generalize on more complex tasks mentioned above.

\subsection{Parameter settings for the experiment in Section \ref{subsect_space_exp}}
\label{supp:sec:param}
The parameters $u_i$'s, the initial values for $\beta_n$'s and $v_n$'s can be treated as hyperparameters. At the beginning of our exploration we randomly choose these initial values. Bit by bit we began to find certain patterns to achieve good performance: for $u_i$'s, usually symmetric; for $v_n$'s, maybe slightly shift towards the negative direction. These are based on tunings and also the observation of the full-precision distribution of weights/activations. Table \ref{lossless-BNN-accuracy-table-param} provides the parameter settings for the experiment in Section \ref{subsect_space_exp}. All ABC-Net models in the experiments are trained using SGD with momentum, and the initial learning rate is set to 0.01.
\begin{table*}[!ht]
\caption{Parameter settings for the experiment in Section \ref{subsect_space_exp}. ``res18", ``res34" and ``res50" are short for Resnet-18, Resnet-34 and Resnet-50 network topology respectively. $M$ and $N$ refer to the number of weight bases and activations respectively. 
}
\label{lossless-BNN-accuracy-table-param}
\begin{center}
\begin{tabular}{c|c|c|c|c|c}
\hline 
\multicolumn{1}{c|}{Network} &\multicolumn{1}{c|}{$M$} &\multicolumn{1}{c|}{$N$} &\multicolumn{1}{c|}{shift parameters ($u_i$'s)} &\multicolumn{1}{c|}{shift parameters ($v_i$'s)} &\multicolumn{1}{c}{$\beta$'s} 
\\ \hline \hline 
res18 & {1} &  1   & 0 & 0.0  &  1.0 \\ 
\hline \hline 
res18 & {3} &  1   & -1,0,1 & 0.0 & 1.0 \\
res18 & {3} &  3   & -1,0,1 & -1.5, 0.0, 1.5 & 1.0, 1.0, 1.0 \\
res18 & {3} &  5   & -1,0,1 & -3.5, -2.5, -1.5, 0.0, 2.5 & 1.0, 1.0, 1.0, 1.0, 1.0\\
\hline \hline 
res18 & {5} &  1   & -2,-1,0,1,2   & 0.0 & 1.0  \\
res18 & {5} &  3   & -2,-1,0,1,2   & -0.9, 0.0 0.9 & 1.0,1.0,1.0 \\
res18 & {5} &  5   & -1,-0.5,0,0.5,1  & -3.5, -2.5, -1.5, 0.0, 2.5 & 1.0, 1.0, 1.0, 1.0, 1.0\\
\hline\hline
res34 & {3} &  3   & -1,0,1 & -3.0, 0.0, 3.0 & 1.0, 1.0, 1.0 \\
\hline\hline
res34 & {5} &  5   & -1,-0.5,0,0.5,1 & -3.5, -2.5, -1.5, 0.0, 2.5 & 1.0, 1.0, 1.0, 1.0, 1.0 \\
\hline\hline
res50 & {5} &  5   & -1,-0.5,0,0.5,1 & -3.5, -2.5, -1.5, 0.0, 2.5 & 1.0, 1.0, 1.0, 1.0, 1.0 \\
\hline
\end{tabular}
\end{center}
\end{table*}

\subsection{Relationship between accuracy and number of binary weight bases $M$}
\label{supp:sec:bwn_acc}
\begin{figure*}[htb]
\centering
\includegraphics[scale = 0.72]{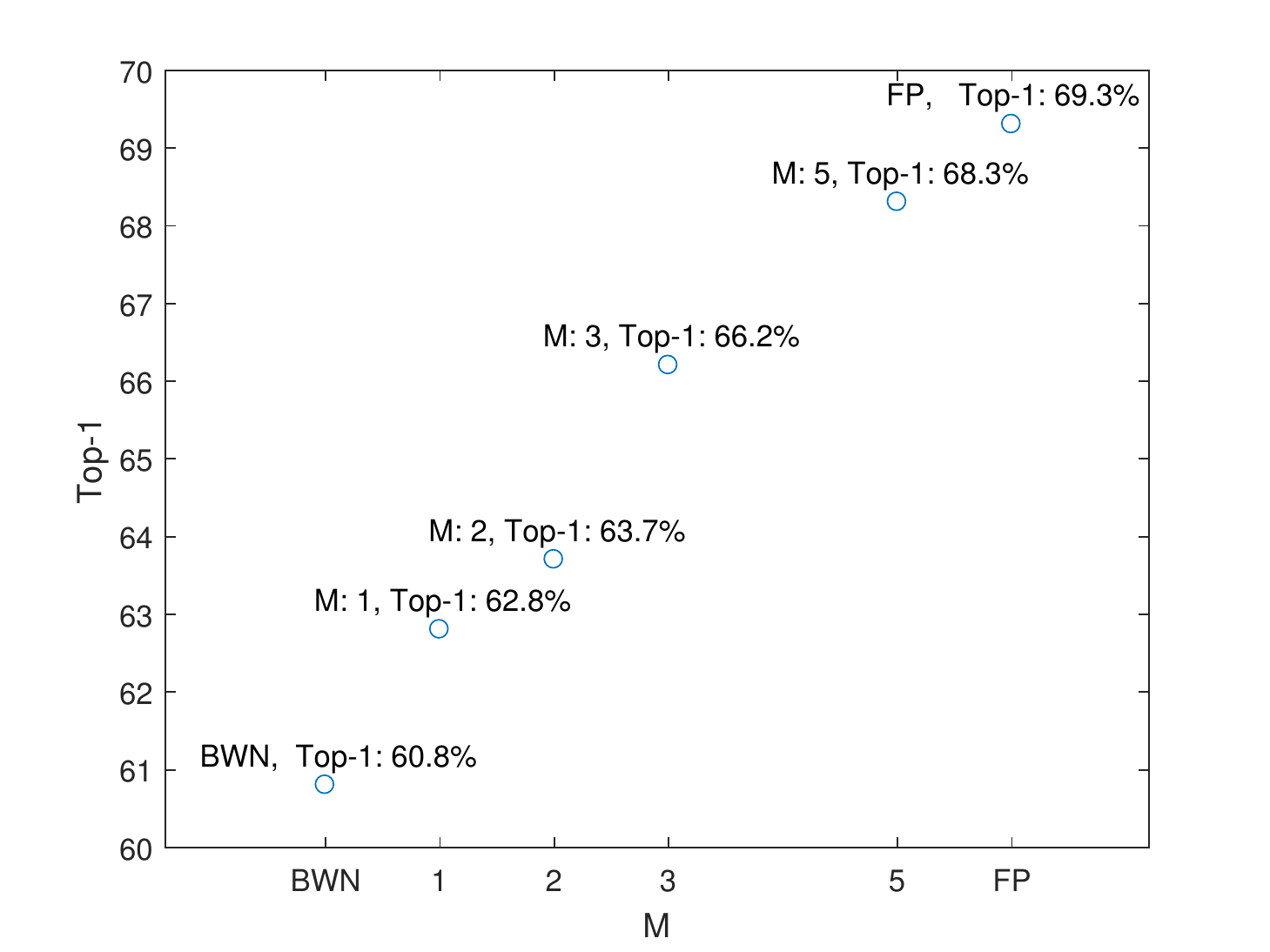}
\\
\includegraphics[scale = 0.72]{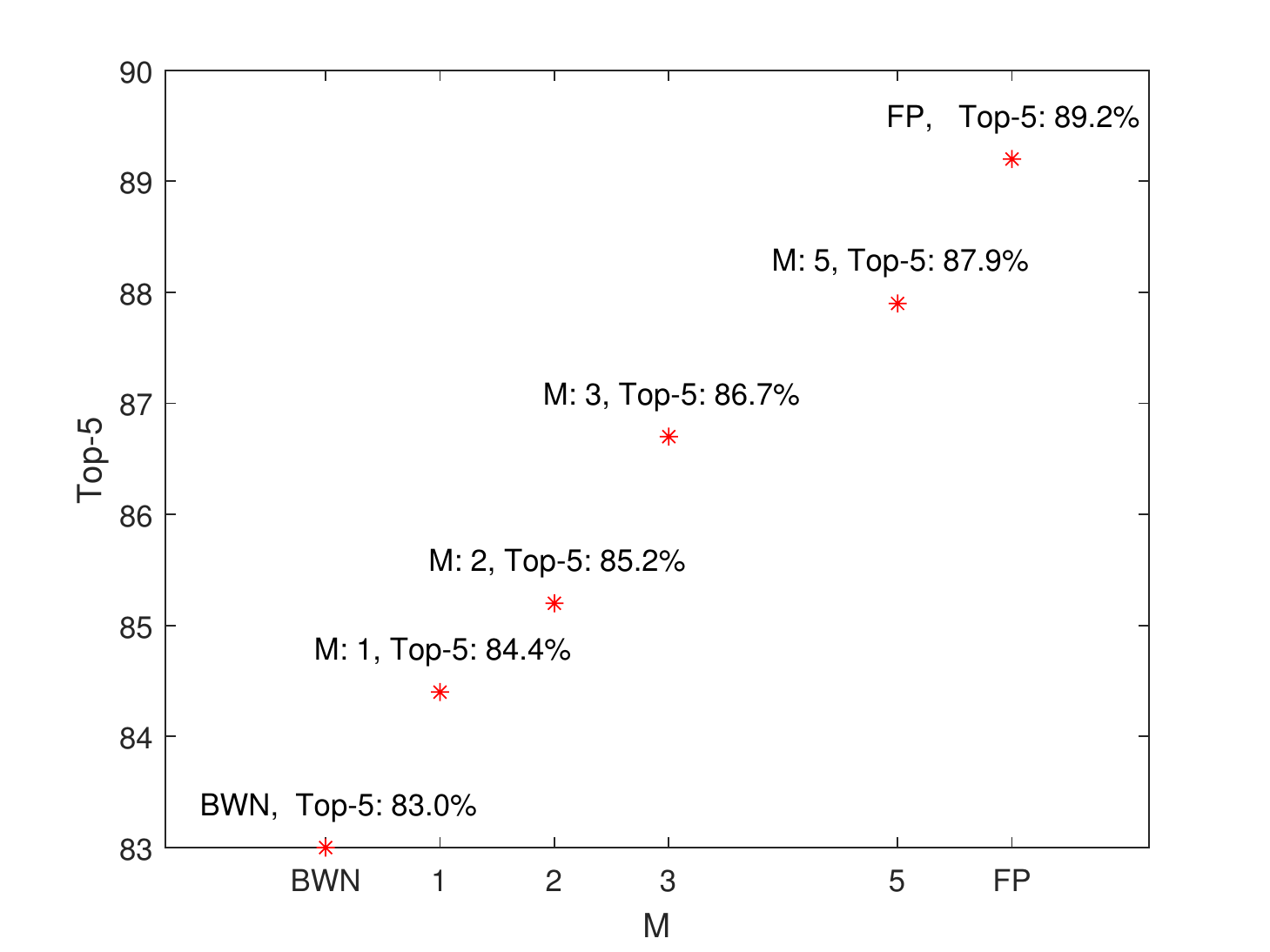}
\caption{
Top-1 (left) and Top-5 (right) accuracy of ABC-Net on ImageNet, using full-precision activation and different choices of the number of binary weight bases $M$.}
\label{fig_bwn_acc}
\end{figure*}

Figure \ref{fig_bwn_acc} shows that the relationship between accuracy and the number of binary weight bases $M$ appears to be linear. Note that we keep the activations being full-precision in this experiment.


\subsection{Application on visual perception of forest trails}
\label{sec:supp:forest}

\subsubsection{VGG-like Network Topology}
\begin{figure*}[htb]
\centering
\includegraphics[scale = 0.9]{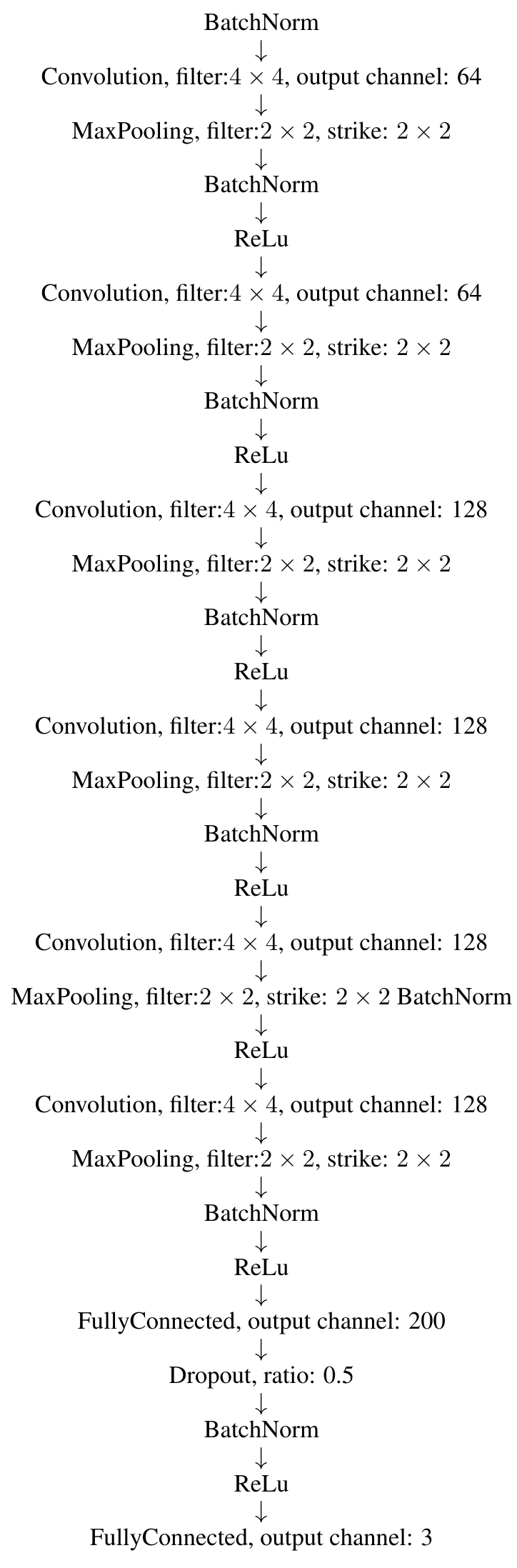}
\caption{The network topology for visual perception of forest trails.}
\label{fig_forestnet}
\end{figure*}

A VGG-like network topology is used for visual perception of forest trails as illustrated in Figure \ref{fig_forestnet}.

\subsubsection{Experiment results on visual perception of forest trails dataset}
\citet{giusti2016machine} cast the forest or mountain trails perception problem for mobile robots as a image classification task based on Deep Neural Networks.
The dataset is composed by 8 hours of $1920\times 1080$ 30fps video acquired by a hiker equipped with three head-mounted cameras . Each image is labelled  in one of three classes: turn right, go straight, turn left. We evaluate ABC-Net against its full precision counterpart using this dataset. The classification result is shown in Table~\ref{table-forest-trails} by fixing both number of weight bases $M$ and activation bases $N$ to be 5. 
{\small
\begin{table*} 
\caption{Classification accuracy on Forest Trails dataset. `FP' stands for `Full Precision'.}
\label{table-forest-trails}
\begin{center}
\begin{tabular}{c|c|c|c|c|c}
\hline 
\multicolumn{1}{c|}{Network}  &\multicolumn{1}{c|}{shift parameters ($u_i$'s)} &\multicolumn{1}{c|}{shift parameters ($v_i$'s)} &\multicolumn{1}{c|}{$\beta$'s} &\multicolumn{1}{c|}{ABC-Net} &\multicolumn{1}{c}{FP} 
\\ \hline \hline 
VGG-like  & -1,0.5,0.0,0.5,1 & 0,0,0,0,0  &  1,1,1,1,1 & 78.0\% & 77.7\%  \\ 
\hline
\end{tabular}
\end{center}
\end{table*}
}
\vspace*{-0.7 cm}

\end{document}